\documentclass{article}

\usepackage{amssymb}
\usepackage{arxiv} 
\usepackage{graphicx}
\usepackage{subcaption}
\usepackage{pdflscape} 
\usepackage{longtable} 
\usepackage{booktabs}  
\usepackage{array}     
\usepackage{enumitem}  
\usepackage{hyperref}       
\usepackage{url}            
\usepackage{amsfonts}       
\usepackage{nicefrac}       
\usepackage[protrusion=true,expansion=false]{microtype} 
\usepackage{lipsum}		
\usepackage{natbib}
\usepackage{doi}

\usepackage{fontspec} 
\usepackage[bidi=default]{babel} 
\babelprovide[main, import]{english}
\babelprovide[import]{persian}

\babelfont[persian]{rm}[Path=./, Extension=.ttf]{Vazirmatn-Light}

\title{Artificial Intelligence for Sentiment Analysis of Persian Poetry}

\author{ \href{https://orcid.org/0000-0003-4507-0822}{\includegraphics[scale=0.06]{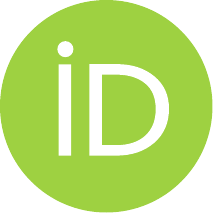}\hspace{1mm}Arash Zargar} \\
	Department of Mechanical \& Industrial Engineering\\ 
    Vector Institute for Artificial Intelligence\\
	University of Toronto\\
	Toronto, Canada \\
	\texttt{a.zargar@mail.utoronto.ca} \\
    \And
    Abolfazl Moshiri \\
	Department of Near \& Middle Eastern Civilizations\\
	University of Toronto\\
	Toronto, Canada \\
	\texttt{a.moshiri@utoronto.ca} \\
    \And
    Mitra Shafaei\\
    Independent Researcher      
    \\
    \And
    Shabnam Rahimi-Golkhandan \\
	Department of Near \& Middle Eastern Civilizations\\
	University of Toronto\\
	Toronto, Canada \\
    \And
    Mohamad Tavakoli-Targhi \\
	Department of Near \& Middle Eastern Civilizations\\
	University of Toronto\\
	Toronto, Canada \\
    \And
    \href{https://orcid.org/0000-0001-5616-8660}{\includegraphics[scale=0.06]{orcid.pdf}\hspace{1mm}Farzad Khalvati$^*$} \\
	Department of Medical Imaging and Institute of Medical Science\\
    Department of Mechanical \& Industrial Engineering \\
    Department of Computer Science \\
    Vector Institute for Artificial Intelligence\\
	University of Toronto\\
	Toronto, Canada \\
	\texttt{farzad.khalvati@utoronto.ca} \\
}

\renewcommand{\shorttitle}{Sentiment Analysis of Persian Poetry}

\begin{document}
\maketitle

\begin{abstract}
Recent advancements of the Artificial Intelligence (AI) have led to the development of large language models (LLMs) that are capable of understanding, analysing, and creating textual data. These language models open a significant opportunity in analyzing the literature and more specifically poetry. In the present work, we employ multiple Bidirectional encoder representations from transformers (BERT) and Generative Pre-trained Transformer (GPT) based language models to analyze the works of two prominent Persian poets: Jal\=al al-D\=in Muḥammad R\=um\=i (Rumi) and Parvin E'tesami. The main objective of this research is to investigate the capability of the modern language models in grasping complexities of the Persian poetry and explore potential correlations between the poems' sentiment and their meters. Our findings in this study indicates that GPT4o language model can reliably be used in analysis of Persian poetry. Furthermore, the results of our sentiment analysis revealed that in general, Rumi’s poems express happier sentiments compared to Parvin E'tesami’s poems. Furthermore, comparing the utilization of poetic meters highlighted Rumi’s poems superiority in using meters to express a wider variety of sentiments. These findings are significant as they confirm that LLMs can be effectively applied in conducting computer-based semantic studies, where human interpretations are not required, and thereby significantly reducing potential biases in the analysis.
\end{abstract}


\section{Introduction}
Recent advancements of Artificial Intelligence (AI) have been such extensive that Geoffrey Hinton, one of the most cited scholars in this fields, compared its effect on the human history to the invention of fire, wheel, and industrial revolution \citep{1-acres2023skynews}. These recent advancements provided extremely powerful methods and tools for textual data analysis through invention of Large Language Models (LLM). An LLM is an advanced deep learning algorithm capable of handling numerous tasks related to natural language processing (NLP). These language models utilize transformer architectures and are developed by training on extensive datasets. Such training allows LLMs to understand, translate, anticipate, or produce text and other forms of content \citep{2-elastic_llms}. Scalability, multilinguality, and efficiency are characteristics of LLMs that make them highly powerful tools specifically for analyzing large textual datasets in the field of literacy and poem analysis. 

In the context of literature, AI-driven tools have shown extensive utility. Text classification is one of these successful applications. In poetry, however, text classification presents a greater challenge compared to other forms of written texts, due to its inherent linguistic complexities \citep{3-rahgozar2020automatic}. Nevertheless, AI tools have previously applied to distinguish poems from prose \citep{4-de2021pattern}, identifying poems’ rhythm \citep{5-agirrezabal2016machine}, style \citep{6-yang2024reimagining}, and chronology \citep{3-rahgozar2020automatic}. \cite{5-agirrezabal2016machine} trained multiple machine learning models, including independent classifiers such as Naive Bayes and sequence labeling models like Hidden Markov model, to extract the rhythm of English poems automatically. Their best model achieved a per-syllable accuracy of 91.4\% and a 55.3\% accuracy in correctly scanning poetry lines, highlighting the need for improving automated rhythm extraction tools through more advanced methods, such as deep learning-based tools. More recently, \cite{6-yang2024reimagining} applied machine learning algorithms to automatically classify modern French poems into the stylistic and thematic subcategories of Romanticism, Parnassism, and Symbolism. They could achieve maximum accuracy of $\sim75\%$ with Support Vector Machine (SVM) algorithm. The authors emphasized the need for more advanced methods to achieve higher levels of accuracy. Similar studies for categorizing on Chinese \citep{7-zhu2023artificial} and Spanish \citep{8-deng2023artificial, 9-navarro2018poetic} poetry based on style and Punjabi \citep{10-kaur2020designing} and Arabic \citep{11-ahmed2019classification} poetry based on subject have been performed previously. In another noteworthy study, \cite{3-rahgozar2020automatic} employed machine learning algorithms to semantically categorize the poetry of Kh\=ajeh Shams-od-D\={\i}n Mo\d{h}ammad \d{H}\=afe\d{z}-e Sh\={\i}r\=az\={\i} (Hafez), one of the greatest Persian poets, predict the chronological order of his works, and tackle the challenge of identifying the original texts from among 16 historical versions of Hafez's Divan.

While artificial intelligence has made significant improvements in classifying and analyzing existing poetic works, AI is increasingly being used to generate original verse. Several studies have explored training AI systems on vast, multilingual literary datasets to compose new poems. \cite{12-kobis2021artificial} and \cite{13-hitsuwari2023does} demonstrated that AI-powered system can generate poems that human participants were unable to distinguish from those composed by humans. Beyond text generation, AI has also demonstrated powerful applications in the conceptual understanding of poetry, including the identification of poetic devices such as metaphors. \cite{14-tanasescu2018metaphor} showed that deep learning outperformed other types of machine learning models in detecting metaphor. 

Another type of analysis that has been the focus of some of the previous studies, is extracting the sentiment of poems using AI. Sentiment analysis of textual data can be considered a classical task that has been the focus of many previous studies in this field \citep{3-rahgozar2020automatic, 15-medhat2014sentiment}. Extracting the sentiment of poems, however, is considered a more difficult task because of the presence of metaphors, amphibologies, and other language-related complexities that require subtle understanding beyond standard sentiment analysis techniques. To this effect, \cite{16-ahmad2020classification} utilized attention-based C-BiLSTM (Convolutional Neural Networks (CNN) and the Bidirectional Long Short-Term Memory (BiLSTM)) models to classify the emotional state of the poems into predefined human emotions such as love, fear, and loneliness. The authors trained their model using 9142 English poems manually labelled with emotional state. Their goal was to develop a computational tool that can predict the emotional state of new poems outside the training dataset and could achieve the accuracy of 88\% in their best performing model. \cite{17-alsharif2013emotion} utilized four supervised learning algorithms including Na\"ive Bayes, SVM, Hyperpipes and Voting Feature Intervals to extract emotions from Arabic poems. Their dataset included 1231 poems gathered from online publicly available sources which contained the main Arabic poem categories of Retha, Ghazal, Fakhr, and Heja. Their analysis concluded that the Hyperpipes algorithm achieved the highest precision of 79\% using non-stemmed, non-rooted, mutually deducted feature vectors with 2000 features. In another study, \cite{18-rajan2020sentiment} utilized a Na\"ive Bayes classifier to determine the sentiment (positive, negative, or neutral) of poems in the Konkani language. These poems were annotated by three native Konkani speakers (two of whom are well-versed in poetry) and the inter-annotator agreement was assessed using the Kappa statistic. Their analysis showed that when the model was tested on poems containing only words with a senti-tag in the Konkani language corpus, the prediction accuracy reached 82\%. However, testing the model on randomly selected poems without ensuring the presence of these words in the corpus resulted in 70\% accuracy. These results highlight the potential for further advancements in sentiment analysis for regional languages using NLP algorithms. In a more recent study, \cite{19-luo2023utilizing} showed that advanced language models like GPT-4 can unveil deeper understandings of English literature by capturing surface-level imagery and uncovering underlying themes such as national pride and reflections on freedom. 

To the best of authors knowledge, no prior research has applied modern Large Language Models (LLMs), such as BERT and GPT, to conduct sentiment analysis on classical Persian poetry. Therefore, this study leverages these advanced models to develop a framework capable of analyzing the sentiment of Persian poetic texts. The primary objective is to map potential correlations between a poem's sentiment and its metrical structure (vazn). To achieve this, we tested BERT and GPT-based models’ capability in determining the sentiment of poems from one of Jal\=al al-D\=in Muḥammad R\=um\=i’s (Rumi) books, Divan-i Shams, as well as the work of a more contemporary Persian poet, Parvin E'tesami. We chose Rumi’s work because Divan-i Shams, with its more than 50 meters, is one of the richest sources in Persian literature for utilizing meters, and includes 13 meters invented by Rumi. Parvin E'tesami’s work was chosen to test the models' performance on more modern texts, allowing us to evaluate their capabilities across a broad temporal spectrum, ranging from the classical medieval era to the twentieth century. Additionally, we wanted to explore how different meters were used to convey various sentiments by a female poet. Considering all these points, Parvin E'tesami's work was an ideal choice as she employed classical meters in her poetry.

\section{Methodology}
\textbf{i.	Dataset Preparation:} This study is conducted using the poems of Divan-i Shams by Rumi and Divan-i Ashaar by Parvin E'tesami. These poems were sourced from the Ganjoor online repository (www.ganjoor.net), a publicly available database of classical Persian poetry. The dataset included each poem's title, verses, and its corresponding poetic meter (Vazn). 

\textbf{ii.	Sentiment Scoring System:} The emotional tone of each poem was evaluated using numerical sentiment scores ranging from 1 to 5, where:
\begin{itemize}
    \item 1 represents a sad or highly negative sentiment.
    \item 2 represents a somewhat sad or mildly negative sentiment.
    \item 3 represents a neutral sentiment.
    \item 4 represents a somewhat happy or mildly positive sentiment.
    \item 5 represents a happy or highly positive sentiment.
\end{itemize}

\textbf{iii.	Preprocessing:}  In natural language processing, a token is the smallest unit of text (e.g., a word or sub-word) that language models use as input for their computations. To facilitate sentiment analysis, all verse lines within a single poem were concatenated into a unified text document and subsequently tokenized. Because the transformer-based models utilized in this study impose maximum input length constraints, poems exceeding these token limits were partitioned into smaller, sequential chunks. The models processed each chunk independently, and the resulting sentiment scores were averaged to compute a global score for the entire poem. Finally, this average was rounded to the nearest integer to align with our 1-to-5 evaluation scale, effectively representing the poem's overall emotional tone.

\textbf{iv.	Language Models:} Four distinct transformer-based language models were employed to perform sentiment analysis on the dataset. Two of the models are based on the encoder-only BERT architecture, while the remaining two utilize the generative GPT architecture. In this study, all models were evaluated in a zero-shot inference setting. For the GPT models, this meant providing prompts without any in-context examples. For the BERT-based models, this involved utilizing versions pre-trained on generic sentiment datasets and applying them directly to our poetry dataset without any domain-specific fine-tuning. These models are outlined below:

\begin{itemize}
    \item \textbf{BERT Multilingual Sentiment Analysis Model:} BERT multilingual model is a pre-trained model that supports 102 languages, and contains 12 layers, 768 hidden units, 12 attention heads, and 110 million parameters \citep{20-devlin2019github, 21-devlin2019bert}. BERT multilingual model is trained on diverse languages and can handle both cased and uncased text. Using the masked language modeling (MLM) approach, BERT was able to learn from both the preceding and succeeding words within a sentence during training. This bidirectional context allowed it to outperform previous models on a wide range of natural language processing tasks. In this study, the BERT Multilingual Uncased model \citep{20-devlin2019github} was used to analyze the sentiment of poems. The model was applied using a tokenizer that split the poems into chunks which fit within the max sequence length of 512 tokens. 
    \item \textbf{Pars-BERT Uncased Model:} While the underlying BERT multilingual model support Persian language and can process Persian text, it was not specifically fine-tunned on Persian sentiment data and therefore, its sentiment predictions may be inaccurate. Thus, we used Pars-BERT model \citep{22-hooshvare_huggingface, 23-farahani2021parsbert}, which is a monolingual fine-tuned version of BERT, specifically trained on Persian language datasets such as Persian Wikipedia (general encyclopedia) containing 1,119,521 documents, Eligasht (itinerary) containing 9,629 documents, Digikala (Digital magazine) containing 8,645 documents, and five other datasets. This domain-specific training data enables Pars-BERT to more effectively capture Persian sentiment nuances compared to generic multilingual models. Performance of Pars-BERT for sentiment analysis in Persian language has been tested by \cite{22-hooshvare_huggingface} on multiple datasets including Digikala (an Iranian e-commerce platform \citep{24-digikala_web}) and SnappFood (an  Iranian online food ordering and delivery service \citep{25-snappfood_web}) which showed a superior results compared to Multilingual BERT. The Digikala user comments dataset, collected by the Open Data Mining Program (ODMP), contains 62,321 user comments distributed across three labels: 10,394 ‘no$\_$idea,’ 15,885 ‘not$\_$recommended,’ and 36,042 ‘recommended \citep{24-digikala_web, 26-digikala_data}.’ Meanwhile, the SnappFood dataset includes 70,000 comments evenly split into ‘Happy’ (35,000) and ‘Sad’ (35,000) \citep{25-snappfood_web, 27-snapfood_data}. Unlike the BERT Multilingual Sentiment analysis model, which classifies the sentiment with a number in 1-to-5 scale, Pars-BERT classifies sentiments as "negative," "neutral," or "positive," which we mapped to the numerical values "1," "3," or "5" to align with the results from other models. Due to Pars-BERT's maximum input sequence length of 512 tokens, we split the longer poems and analyzed their sentiments in separate chunks.
    \item \textbf{GPT 4o-mini and GPT 4o Models:} GPT-4o is one of OpenAI's most advanced multimodal large language models, capable of processing both text and image inputs and generating text outputs \citep{28-openai_gpt4o}. GPT-4o contains a deep transformer architecture that uses cross-attention and multi-head attention to facilitate information exchange among various input modalities. GPT-4o has demonstrated excellent performance in non-English languages which makes it highly suitable for analyzing Persian poetry. This model’s strong multilingual performance suggests it has been trained or fine-tuned on a wide variety of languages, including Persian. GPT-4o-mini \citep{29-openai_gpt4o_mini} is a smaller, more affordable version of GPT-4o, designed for fast, lightweight tasks. In this study both GPT models were accessed through OpenAI's API, and sentiments was assigned using the custom prompt below: 
    \newline Prompt template = \textit{“Analyze the sentiment of the following poem and return a number between 1 and 5, where 1 means sad, 5 means happy, 3 is neutral, 2 and 4 are intermediate cases. RETURN ONLY ONE NUMBER THAT SHOWS THE SENTIMENT, (NO LONG ANSWERS JUST A NUMBER)”}
    
    Using this prompt, the models were tasked with returning a single sentiment score between 1 and 5, similar to the BERT multilingual model. GPT-4o and GPT-4o-mini both have a context window of 128,000 tokens, which allowed us to analyze all the poems in our dataset without splitting the poems. 
\end{itemize}

\textbf{v.	Comparing the sentiment scores with human annotated poems:}
To ensure the accuracy of our language models and validate the sentiment scores generated, we selected 100 poems from Rumi’s Divan-i Shams and had them evaluated by 2 scholars specializing in humanities and 2 other annotators with general knowledge about Farsi literature. Each scholar was asked to rate the sentiment of each poem on a scale of 1 to 5. Importantly, these poems were carefully chosen to represent a wide range of meters found in Divan-i Shams. The sentiment ratings provided by the scholars served as a benchmark to compare with the sentiment scores generated by our language models. This comparison allowed us to identify the model that most accurately captures the emotional tone of the poems. Before generating a Ground Truth label for the 100 chosen poems, we first assessed the consistency of the human annotators using Krippendorff’s Alpha metric to understand the noise level in the dataset. This metric is chosen because it recognizes that a disagreement between 1 and 5 is more severe than a disagreement between 4 and 5. Furthermore, this metric can compare 4 annotator’s results altogether and it is robust for small sample size data sets.  Due to the subjective nature of poetic interpretation, perfect inter-annotator agreement was neither expected nor observed. Consequently, establishing a robust "Ground Truth" (Gold Label) was the primary statistical prerequisite before model evaluation could be performed. To determine the most reliable method for aggregating human responses, we compared four aggregation strategies:

\begin{itemize}
    \item Mean: The arithmetic average of the 4 human grades (rounded to the nearest integer). This is appropriate for minimizing total error distance.
    \item Median: The middle value of the grades. This is statistically robust for ordinal data as it ignores outliers.
    \item Mode: This considers the majority vote. Mode method is specifically accurate for consensus but struggles when there is high variance.
    \item Dawid-Skene (DS) Model: An Expectation-Maximization (EM) algorithm that estimates the "true" label by iteratively modeling the confusion matrix (reliability) of each annotator.
\end{itemize}

We validated these methods by calculating the Average Quadratic Weighted Kappa (QWK) of each aggregation method against the individual human annotators. This metric was chosen because it penalizes disagreements based on the distance between the ratings. For each Ground Truth candidate (e.g., DS, Mean etc.), we calculated the Quadratic Weighted Kappa (QWK) agreement between that Ground Truth and each of the human annotators. Afterwards, we averaged these 4 scores to get a single alignment score. The method with the highest Average Quadratic Weighted Kappa (QWK) was selected as the final Ground Truth. Afterwards, we evaluated four models, i.e., Bert Multilingual, Pars-BERT, GPT-4o-mini, and GPT-4o, against the established Ground Truth by calculating the absolute accuracy and QWK correlation of their predicted sentiments with the Ground Truth labels.

\textbf{vi. Entropy calculation:}
To evaluate the diversity of sentiments expressed within a single poetic meter by each poet, we utilized entropy as a statistical metric. Entropy, denoted as $H(X)$ and calculated using the following equation, quantifies the distribution of sentiment scores associated with a specific metrical structure.

\begin{equation}
        H (X) = -\sum_{i=1}^{n} p(x_i) \log_2p(x_i)
        \label{eq:entropy}
\end{equation}

Where, $H(X)$ is the entropy, $p(x_i)$ is the probability of outcome $x_i$, and $n$ is the number of possible outcomes.

\textbf{vii. Investigating the Correlation Between Poetic Meters and Sentiment in Persian Poetry:}
One of the primary ideas of this research was to explore whether there is a correlation between sentiments of poems in Persian language with meters used. This has been achieved by performing different statistical analysis and several parameters, such as average sentiment score, entropy, standard deviation, and polarization on the calculated sentiments on Rumi’s Divan-i Shams and Parvin E'tesami’s book.

\textbf{viii. Naming Convention for Poetic Meters:}
To prevent visual clutter and simplify the graphical representation of our results, we established an alphanumeric naming convention for the poetic meters. As detailed in Table \ref{tab:1-meterMapping}, each meter is designated by a prefix letter followed by a numerical index. Meters shared by both Rumi and Parvin E'tesami are denoted with the prefix 'C' (Common). Conversely, meters exclusive to Rumi’s or Parvin E'tesami’s works are prefixed with 'R' and 'P,' respectively. This systematic approach ensures optimal clarity and readability across all subsequent figures.

\clearpage 
\begin{landscape}

\begin{table}
  \caption{Mapping of poetic meters.}
  \label{tab:1-meterMapping}
  \begin{tabular}{c}
    \includegraphics[width=20cm]{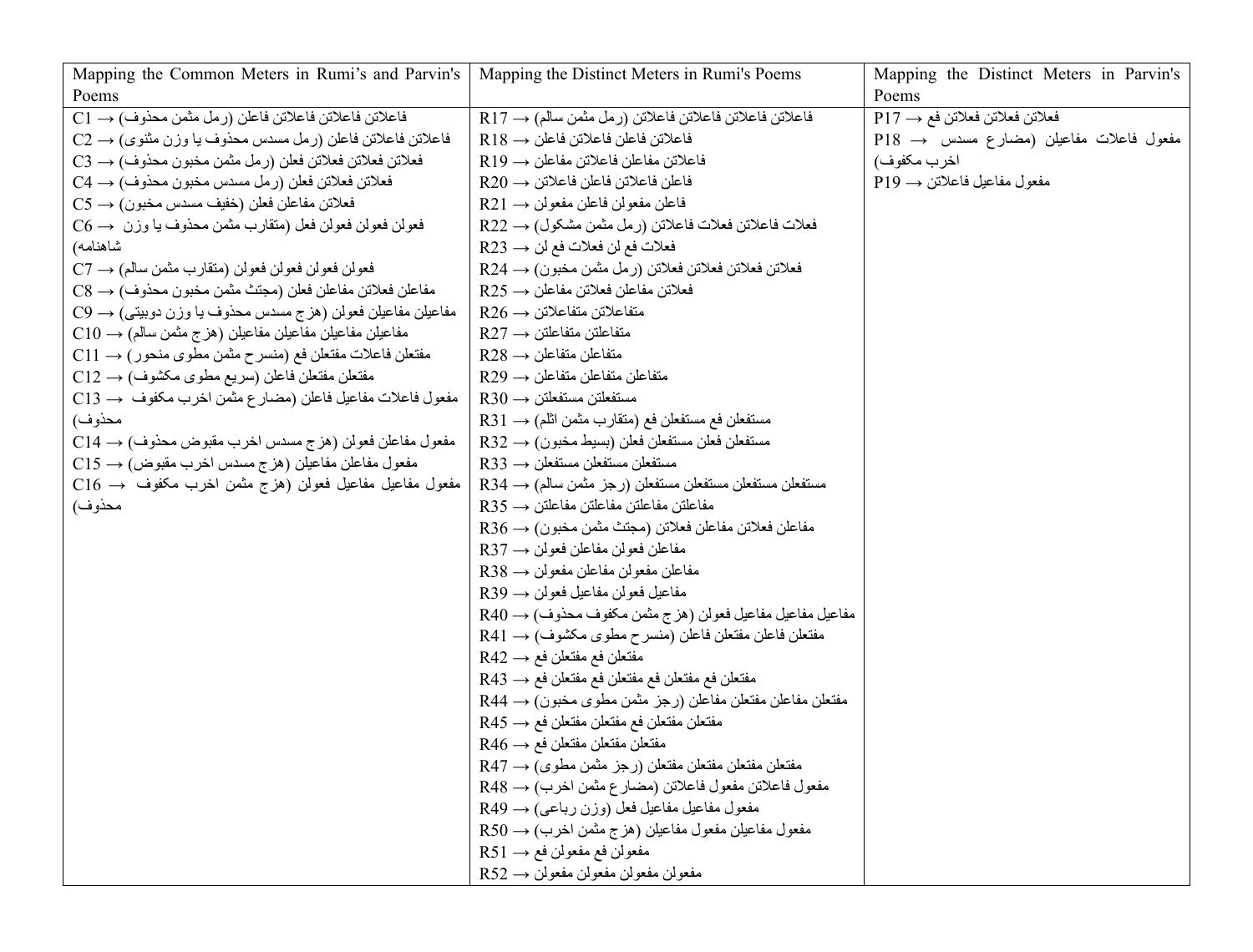}\\
\end{tabular}
\end{table}
\end{landscape}
\clearpage

\section{Results}

Table \ref{tab:2-fleiss_kappa} presents the Nominal Fleiss’ Kappa parameter, calculated from sentiment scores assigned to each poem by the language models. This parameter reflects the repeatability of the data and serves as a measure of the reliability of the analysis. The Nominal Fleiss’ Kappa is computed by performing sentiment analysis six times for Rumi’s poems and three times for Parvin E'tesami’s poems.

\begin{table}[htbp]
    \centering
    \caption{Nominal Fleiss' Kappa parameter based on the seiment scores of vaious language models.}
    \label{tab:2-fleiss_kappa}
    \begin{tabular}{l c c c c}
        \toprule
        & \textbf{BERT Multilingual} & \textbf{Pars-Bert} & \textbf{GPT 4o-mini} & \textbf{GPT 4o} \\
        \midrule
        \textbf{Rumi} & 1.0 & 1.0 & 0.9481 & 0.9578 \\
        \textbf{Parvin} & 1.0 & 1.0 & 0.9882 & 0.9316 \\
        \bottomrule
    \end{tabular}
\end{table}

\begin{figure}[h]
  \centering
  \includegraphics[width=\linewidth]{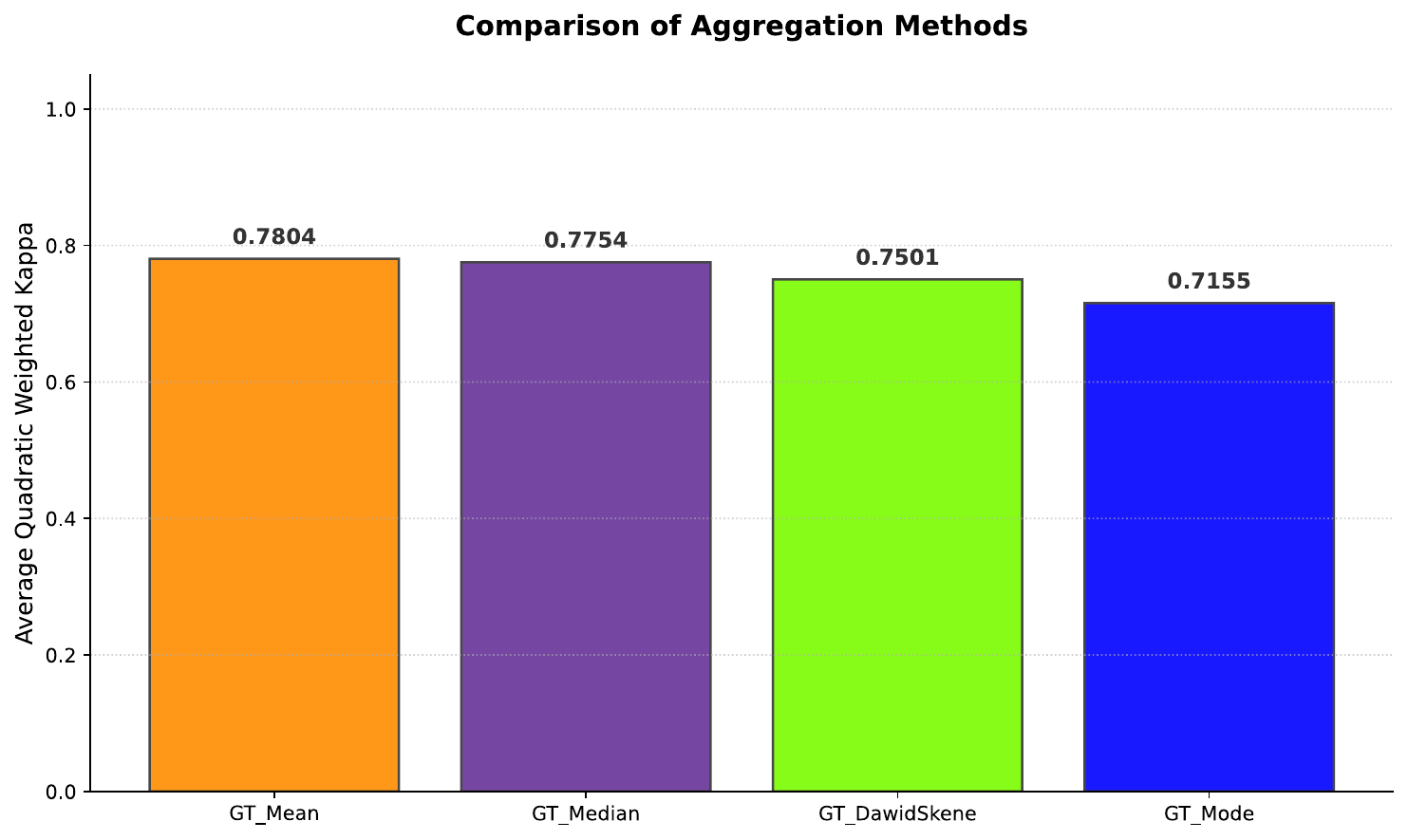}
  \caption{Average Quadratic Weighted Kappa (QWK) of each aggregation method against the individual human annotators. Mean method shows the best performance.}
  \label{fig:1-qwk_aggregation}
\end{figure}

To evaluate the reliability of our human annotations, we calculated Krippendorff’s Alpha for the sampled Rumi dataset. The resulting value of 0.6 indicates a moderate level of agreement among the annotators. To determine the most accurate ground truth labels for the 100 selected poems, we implemented and compared four aggregation strategies: one probabilistic method (Dawid-Skene (DS)) and three heuristic methods (Mean, Median, and Mode). Figure \ref{fig:1-qwk_aggregation} illustrates the Average Quadratic Weighted Kappa (QWK) of each aggregation method compared against the individual human annotators.

\begin{figure}[htbp]
    \centering
    \begin{subfigure}[b]{0.48\textwidth}
        \centering
        \includegraphics[width=\textwidth]{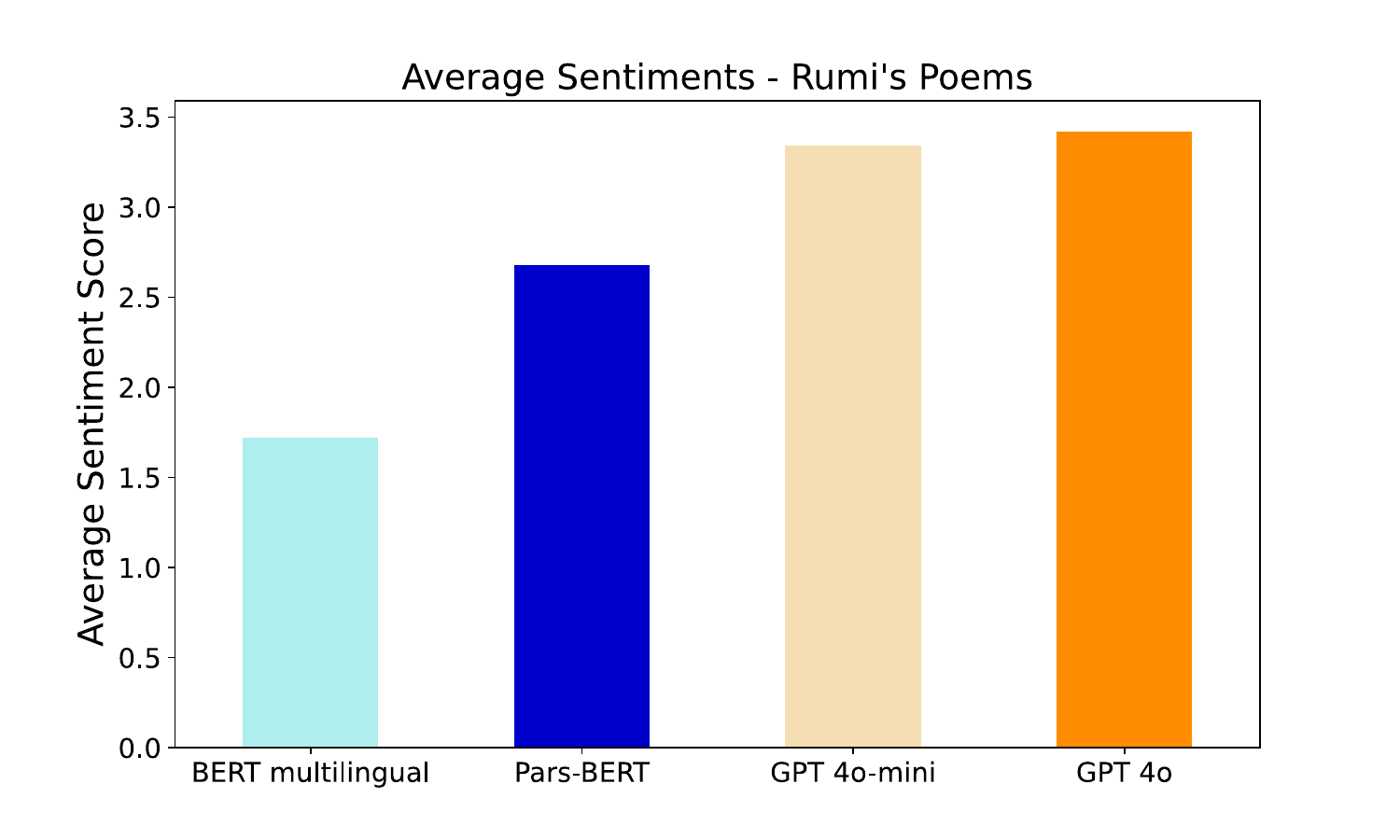}
        \caption{}
        \label{fig:2a}
    \end{subfigure}
    \hfill
    \begin{subfigure}[b]{0.48\textwidth}
        \centering
        \includegraphics[width=\textwidth]{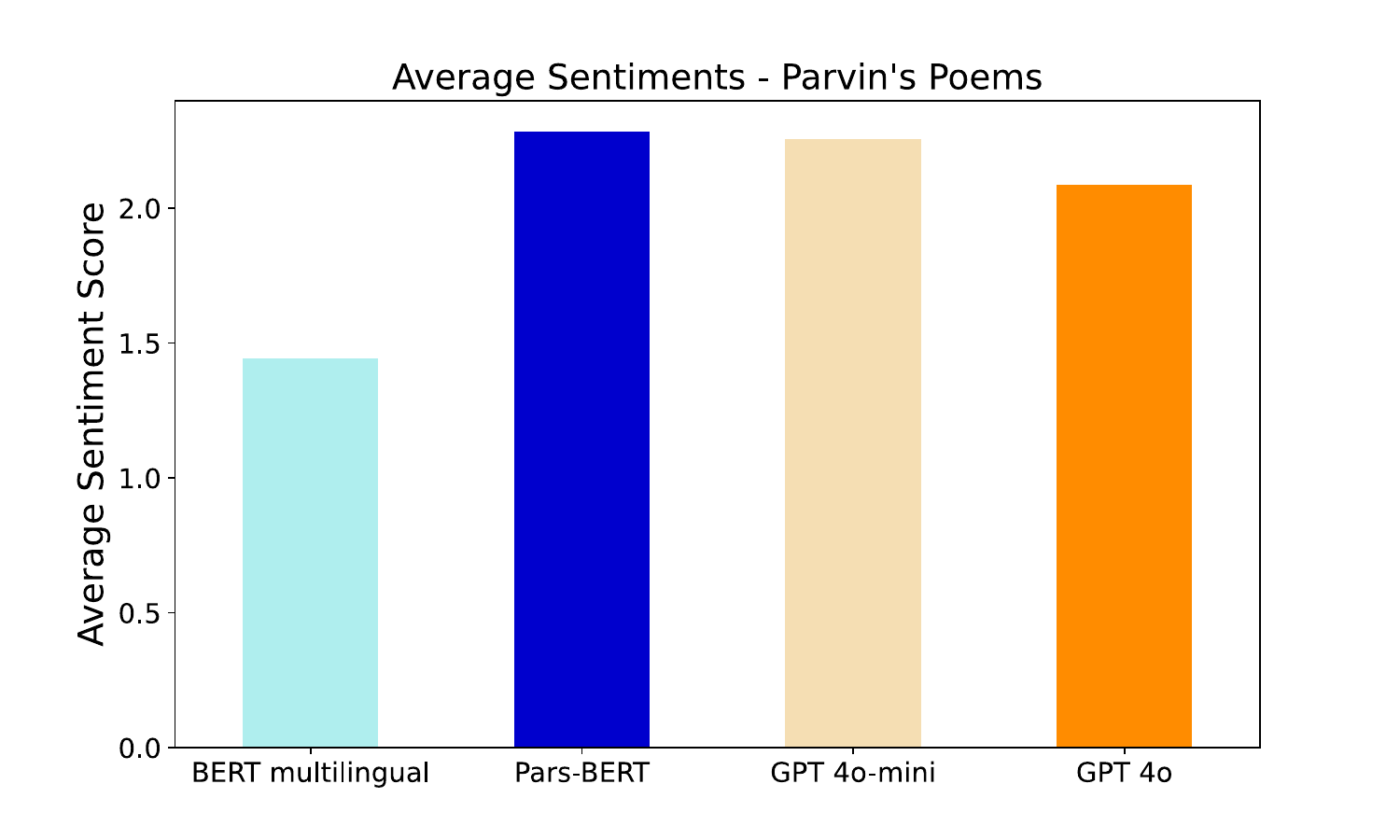}
        \caption{}
        \label{fig:2b}
    \end{subfigure}
    
    \vspace{0.5cm} 
    
    \begin{subfigure}[b]{0.48\textwidth}
        \centering
        \includegraphics[width=\textwidth]{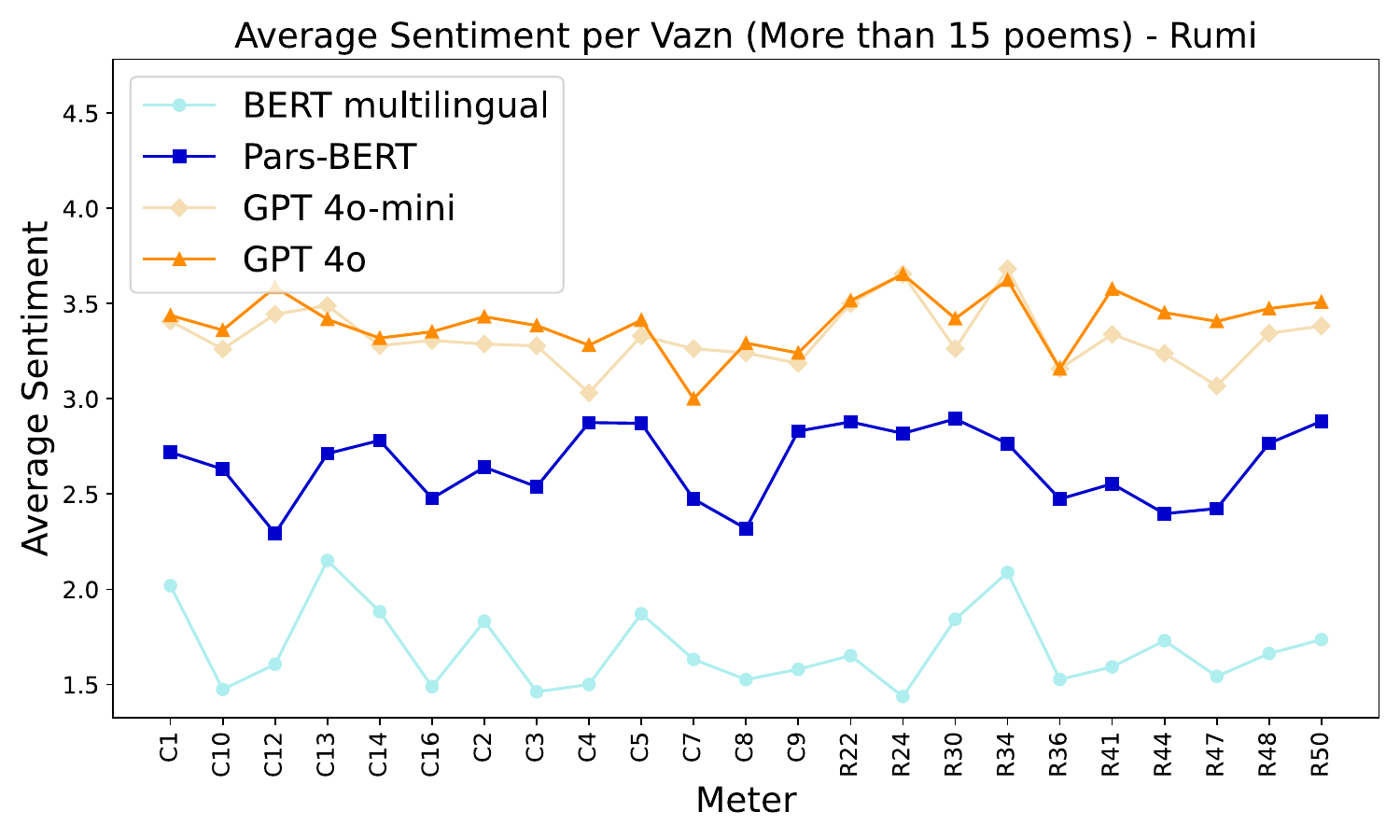}
        \caption{}
        \label{fig:2c}
    \end{subfigure}
    \hfill
    \begin{subfigure}[b]{0.48\textwidth}
        \centering
        \includegraphics[width=\textwidth]{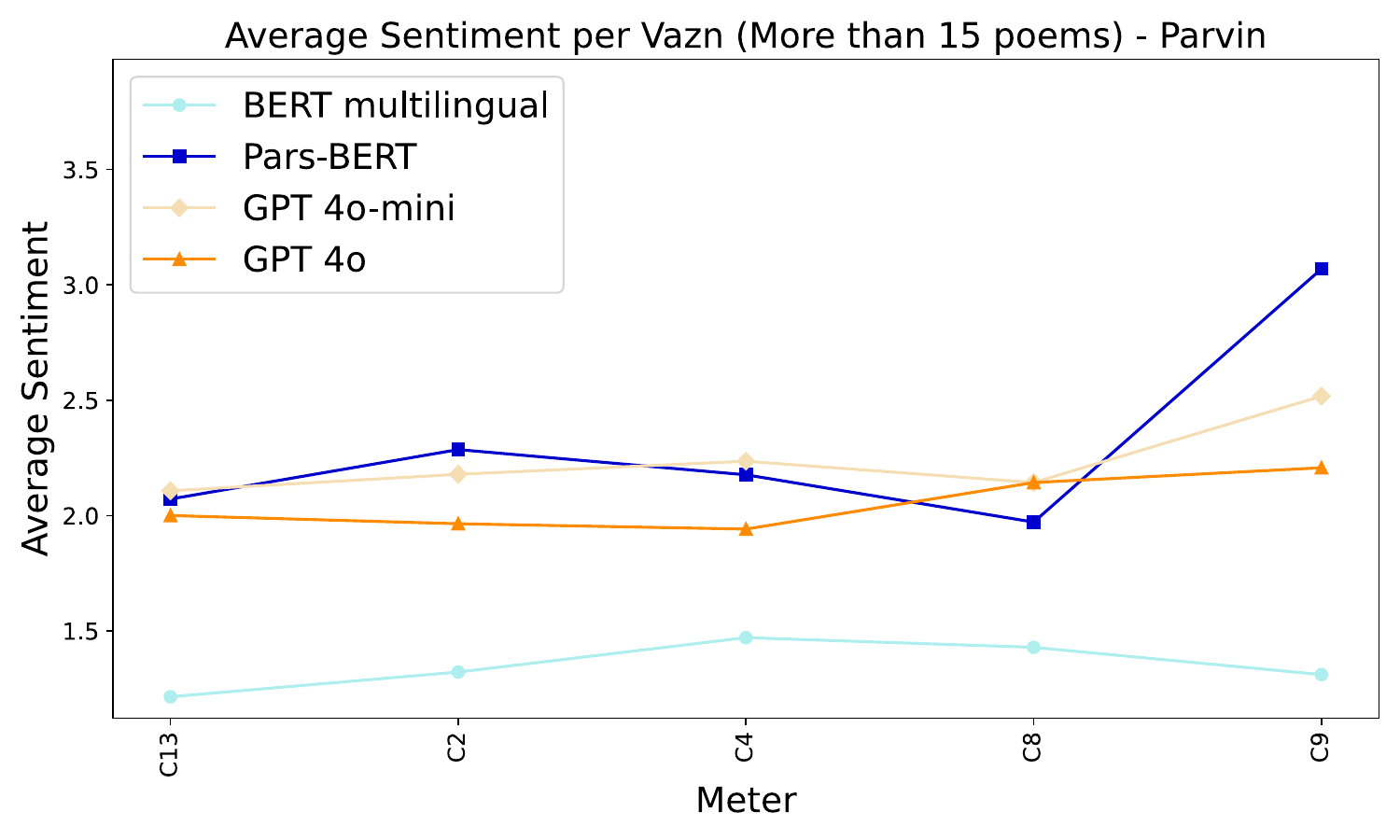}
        \caption{}
        \label{fig:2d}
    \end{subfigure}
    
    \caption{Comparison of average sentiment scores assigned by various LLMs to Rumi’s and Parvin E'tesami’s poems. (a) Average sentiment for Rumi’s poems across LLMs. (b) Average sentiment for Parvin E'tesami’s poems across LLMs.(c) Average sentiment for each meter in Rumi’s poems that contain more than 15 poems across LLMs. (d) Average sentiment for each meter in Parvin E'tesami’s poems that contain more than 15 poems across LLMs. The results highlight significant differences in the average sentiment scores returned by different LLMs. Furthermore, it is observed that there is a consistent trend of higher sentiment scores in Rumi’s work regardless of the model used.}
    \label{fig:2-sentiment_comparison}
\end{figure}

Figures \ref{fig:2a} and \ref{fig:2b} show the average sentiment for each book across the different language models. Figures \ref{fig:2c} and \ref{fig:2d} present the average sentiment scores for the meters that are used for composing at least 15 poems in the corresponding book.

To quantify the alignment between the sentiment analysis models and the established ground truth (Mean aggregation), Table \ref{tab:3-qwk_scores} summarizes the Quadratic Weighted Kappa metrics for all models and human annotators, measured against the ground truth.

\begin{table}[htbp]
    \centering
    \caption{Quadratic Weighted Kappa (QWK) agreement scores and absolute prediction accuracy for human annotators and sentiment analysis models compared against the Ground Truth (Mean aggregation) labels. GPT 4o model shows the highest degree of alignment with human evaluations.}
    \label{tab:3-qwk_scores}
    \resizebox{\textwidth}{!}{%
    \begin{tabular}{l c c c c c c c c}
        \toprule
        \textbf{Sentiment evaluator} & \textbf{Human annotator 1} & \textbf{Human annotator 2} & \textbf{Human annotator 3} & \textbf{Human annotator 4} & \textbf{Pars-Bert} & \textbf{Bert Multilingual} & \textbf{GPT 4o-mini} & \textbf{GPT 4o} \\
        \midrule
        Quadratic Weighted Kappa score & 0.8017 & 0.8025 & 0.7969 & 0.7206 & 0.0006 & 0.0467 & 0.5024 & 0.6003 \\
        Absolute Prediction Accuracy (\%) & 48 & 47 & 61 & 51 & 6 & 7 & 25 & 33 \\
        \bottomrule
    \end{tabular}%
    }
\end{table}

\begin{figure}[htbp]
    \centering
    \begin{subfigure}[b]{0.48\textwidth}
        \centering
        \includegraphics[width=\textwidth]{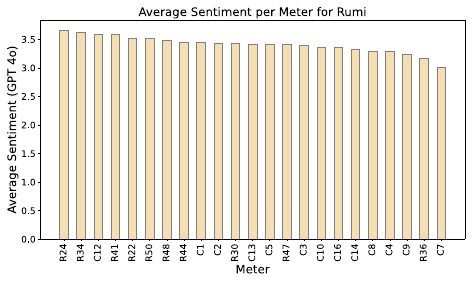}
        \caption{}
        \label{fig:3a}
    \end{subfigure}
    \hfill
    \begin{subfigure}[b]{0.48\textwidth}
        \centering
        \includegraphics[width=\textwidth]{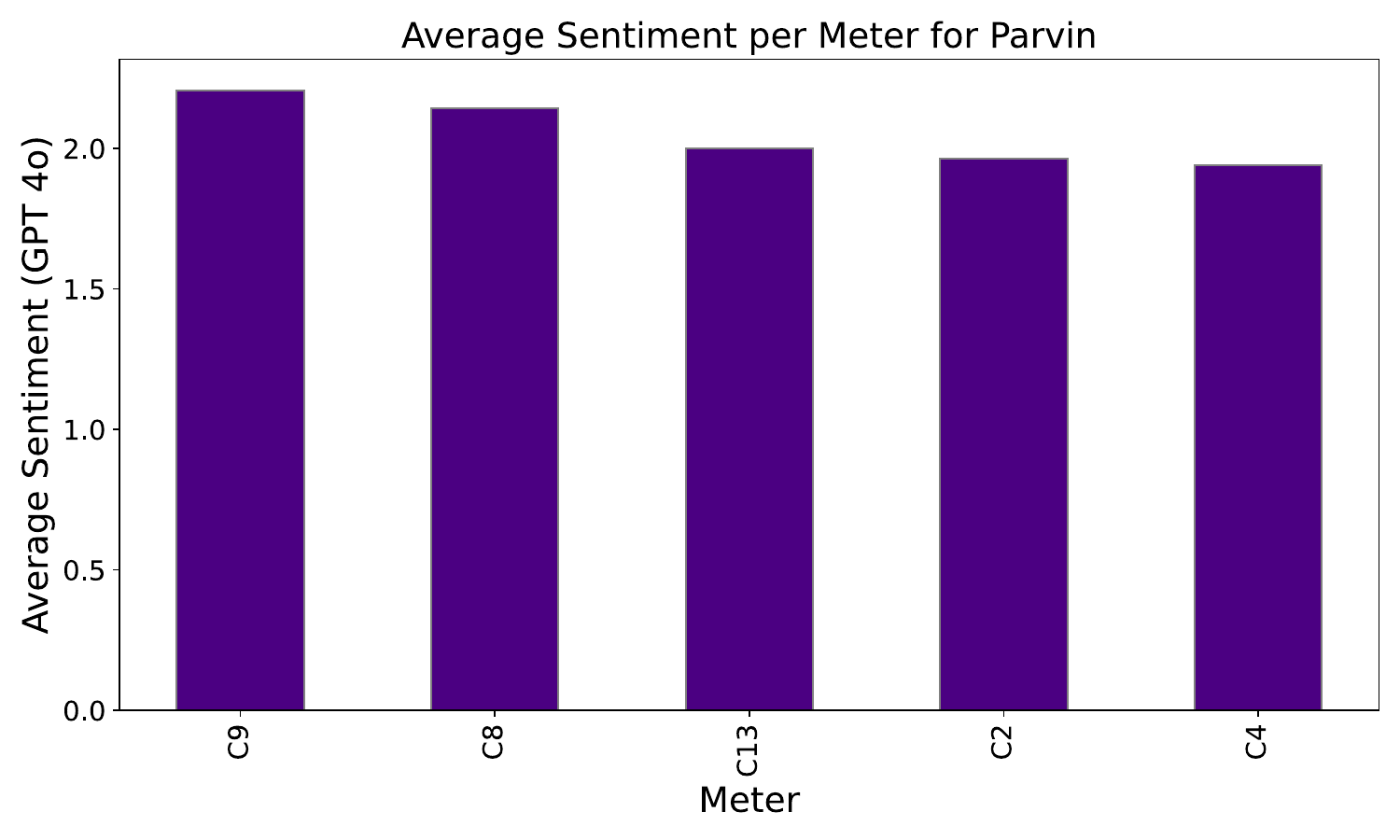}
        \caption{}
        \label{fig:3b}
    \end{subfigure}
    
    \vspace{0.5cm} 
    
    \begin{subfigure}[b]{0.48\textwidth}
        \centering
        \includegraphics[width=\textwidth]{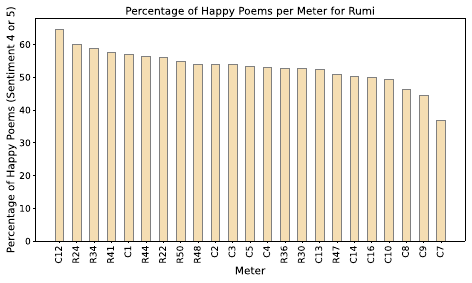}
        \caption{}
        \label{fig:3c}
    \end{subfigure}
    \hfill
    \begin{subfigure}[b]{0.48\textwidth}
        \centering
        \includegraphics[width=\textwidth]{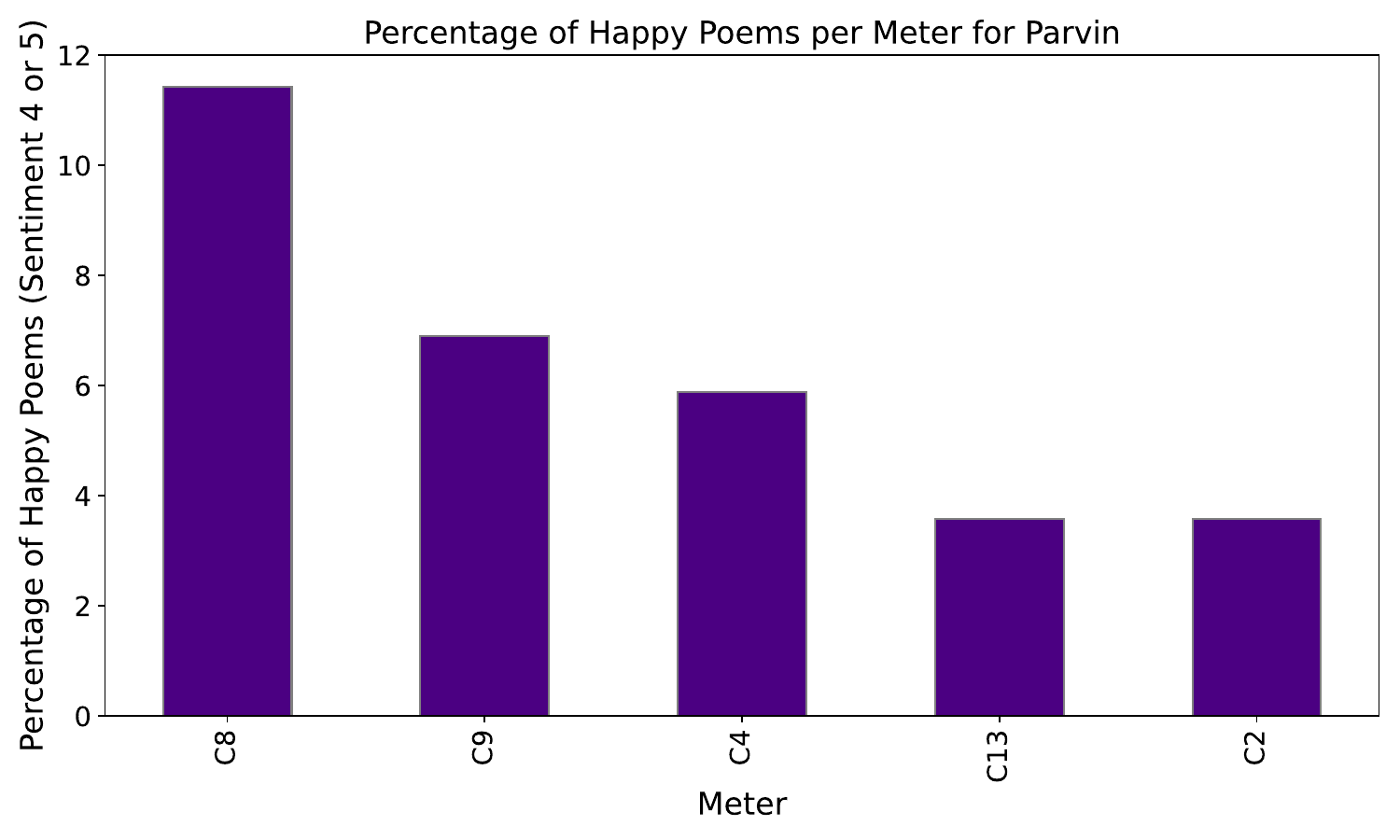}
        \caption{}
        \label{fig:3d}
    \end{subfigure}
    
    \vspace{0.5cm} 
    
    \begin{subfigure}[b]{0.48\textwidth}
        \centering
        \includegraphics[width=\textwidth]{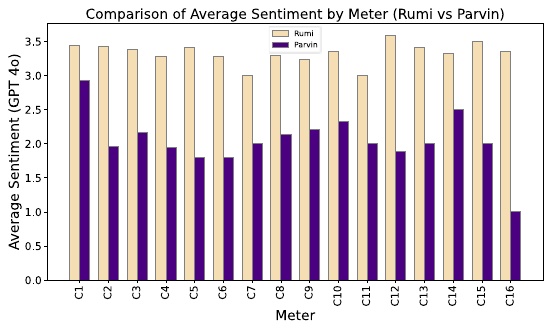}
        \caption{}
        \label{fig:3e}
    \end{subfigure}
    
    \caption{Identification of poetic meters associated with positive sentiment in Rumi’s and Parvin E'tesami’s poems. (a) Meters (containing more than 15 poems) with the highest average sentiment scores in Rumi’s poems. (b) Percentage of Rumi’s poems with happy sentiments in the identified meters (containing more than 15 poems). (c) Meters (containing more than 15 poems) with the highest average sentiment scores in Parvin E'tesami’s poems. (d) Percentage of Parvin E'tesami’s poems with happy sentiments in the identified meters (containing more than 15 poems). (e) Comparison of common meters used by both poets, showing distinction between utilizing poems in the poets’ artworks.}
    \label{fig:3-sentiment_meters}
\end{figure}

For the remainder part of this paper, we have utilized GPT4o model for our analyses, as this model’s results were more aligned with the sentiment annotations provided by our scholar group.  Figures \ref{fig:3a} and \ref{fig:3c} identify the meters with highest mean sentiment scores (happiest) in Rumi’s and Parvin E'tesami’s poems, respectively. Figures \ref{fig:3b} and \ref{fig:3d} depict the meters with the highest percentage of poems exhibiting happy sentiments (sentiment scores of 4 and 5) for Rumi’s and Parvin E'tesami’s poems, respectively. 

\begin{figure}[htbp]
    \centering
    \begin{subfigure}[b]{0.48\textwidth}
        \centering
        \includegraphics[width=\textwidth]{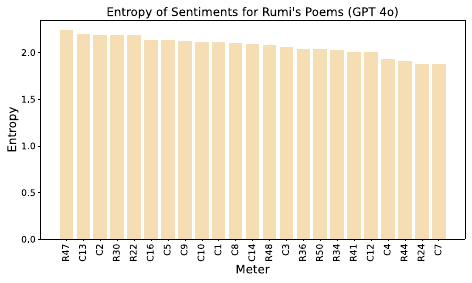}
        \caption{}
        \label{fig:4a}
    \end{subfigure}
    \hfill
    \begin{subfigure}[b]{0.48\textwidth}
        \centering
        \includegraphics[width=\textwidth]{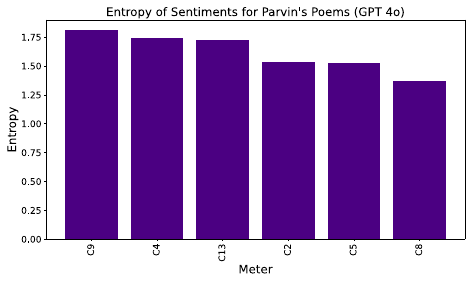}
        \caption{}
        \label{fig:4b}
    \end{subfigure}
    
    \caption{Sentiment entropy across poetic meters in Rumi’s and Parvin E'tesami’s poems. (a) Entropy of sentiment distribution in Rumi’s poems. (b) Entropy of sentiment distribution in Parvin E'tesami’s poems. The results highlight Rumi’s masterful use of meters to convey diverse sentiments.}
    \label{fig:4-entropy_comparison}
\end{figure}

Figure \ref{fig:4-entropy_comparison} illustrates the entropy of meter usage by Rumi (Figure \ref{fig:4a}) and Parvin E'tesami (Figure \ref{fig:4b}). Entropy is a parameter that shows variations of expressed emotions using a specific meter.

\begin{figure}[htbp]
    \centering
    \begin{subfigure}[b]{0.45\textwidth}
        \centering
        \includegraphics[width=\textwidth]{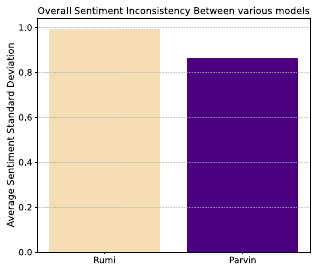}
        \caption{}
        \label{fig:5a}
    \end{subfigure}
    \hfill
    \begin{subfigure}[b]{0.51\textwidth}
        \centering
        \includegraphics[width=\textwidth]{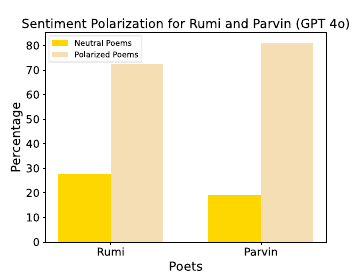}
        \caption{}
        \label{fig:5b}
    \end{subfigure}
    
    \caption{(a) Standard deviation of sentiment scores in Rumi’s and Parvin E'tesami’s poems, illustrating the variability of emotional expression. Rumi’s poems show a wider range of deviations from the average sentiment. (b) Sentiment polarization in Rumi’s and Parvin E'tesami’s poems, distinguishing between neutral and polarized (very sad or very happy) poems.}
    \label{fig:5-std_polarization}
\end{figure}

Figure \ref{fig:5a} shows the standard deviation of the sentiment scores from the average sentiment across all of poems in Divan-i Shams and Parvin E'tesami’s book. Figure \ref{fig:5b} shows the sentiment polarization (scores of 1, 2, 4,or 5) and neutrality (score of 3) in Rumi’s poems, and Parvin E'tesami’s book.

\section{Discussions}

To evaluate the performance of four distinct sentiment analysis models on our dataset of Farsi poems, we curated a dataset of 100 samples selected from Rumi’s poetry. Rumi’s works were chosen specifically for their high density of metaphors and archaic linguistic structures which presents a more rigorous challenge for semantic interpretation compared to more contemporary work of Parvin E'tesami. The Krippendorff’s Alpha for our dataset between our four annotators is equal to 0.6. This indicates "Moderate Agreement," and confirms that the task is highly subjective. Because the humans do not align perfectly, we cannot simply trust one human as the " ground truth" label for 100 selected poems. We were required to perform a statistical aggregation method to filter out noise and bias. As illustrated in \ref{fig:1-qwk_aggregation}, the Mean method demonstrated the highest alignment with the human consensus by achieving an average Quadratic Weighted Kappa (QWK) of $\sim0.78$. Consequently, the labels generated by the Mean model were adopted as the Ground Truth for all subsequent model evaluations.

The first question that may arise about analyzing the sentiment of poems using AI, is that how accurate models’ predictions are, and whether the LLMs can understand the complexities of poetry. To gain insight into the ability of these models in analysing poetic sentiments, we first compared the average sentiment scores for the poems calculated using different LLMs. As shown in Figure \ref{fig:2a} and Figure \ref{fig:2b}, the average sentiment for each book varies significantly across the different language models. Interestingly, however, all language models assign a higher sentiment scores to Rumi’s poems compared to Parvin E'tesami’s. This suggests that regardless of what model we choose for analyzing the sentiments, the LLMs recognize Rumi’s poems conveying happier sentiment compared to Parvin E'tesami’s. We have repeated the same process for each of meters that contain more than 15 poems and observed the same trends between Rumi’s and Parvin E'tesami’s poems. While this is an interesting finding, the question regarding the accuracy of calculated sentiments of Persian poems with LLMs now becomes twofold: 1) how reliable are these predictions overall?, and 2) which model determines the sentiment of poems more accurately? To evaluate reliability of calculated sentients, we performed the analysis multiple times and compared the results by calculating Nominal Fleiss’ Kappa parameter. As shown in Table \ref{tab:1-meterMapping}, Kappa parameter for all models is higher than 0.93 which indicates the repeatability of the results. To evaluate the accuracy of models in determining poems’ sentiments, we must compare their evaluations with human interpretations. Notifying the fact that analysis of sentiment is a subjective task by its nature, and even human interpretations of a single text significantly vary. To verify the accuracy of our sentiment analysis, we selected 100 poems and asked 2 scholars specializing in Persian poetry and 2 annotators with general Farsi literature knowledge, to rate their emotional responses to each poem. Of these 100 poems, 80 are chosen from those with highest standard deviations in sentiment analysis results across different language models, meaning that these are the poems that LLMs significantly disagree on their sentiments. The remaining 20 poems, however, are selected from those that various LLMs predicted sentiments are identical. Our initial hypothesis was that if the scholars' evaluations aligned with the models, these poems could be categorized as having "straightforward" sentiments that are easy to interpret for both humans and machines. However, our analysis revealed a significant discrepancy. Despite the consensus among the LLMs on these 20 poems, we observed no significant correlation among the human annotators, nor between the humans and the models. This lack of alignment indicates that a high degree of model agreement does not necessarily imply that a poem’s sentiment is objectively clear. Furthermore, it suggests that the underlying mechanisms used by LLMs to determine sentiment differ fundamentally from the interpretive processes of human annotators, making it impossible to categorize these poems as having universally "clear" sentiments across both domains.

Given this inherent complexity and the divergence between human and machine interpretation, it became essential to identify which model could best approximate the human consensus, despite the challenges. We benchmarked each model against the established Ground Truth (derived via Mean aggregation). The results in Table \ref{tab:3-qwk_scores} indicate a significant performance gap between the GPT-based LLMs and the BERT-based models. GPT-4o emerged as the top performer with a QWK of 0.6. While the exact accuracy remained low ($\sim33\%$), this figure must be interpreted within the context of the task's subjectivity. Notably, even human annotators, including scholars in the field, did not achieve high exact agreement with the Ground Truth labels. As shown in Table \ref{tab:3-qwk_scores}, human exact accuracy ranged from 47\% to 61\%. This demonstrates that a 33\% accuracy for a machine model is not a failure of computation, but rather a reflection of the inherent ambiguity of the task. When humans themselves only agree on the exact sentiment score roughly half the time, the model’s performance is relatively more impressive than the raw number suggests.

Moreover, Our analysis suggests that while the GPT-based LLMs often miss the exact intensity of the sentiment (e.g., predicting a 4 instead of a 5), they correctly identifies the directional polarity of the poem. Conversely, the smaller encoder-based models (Bert Multilingual and Pars-Bert) struggled to capture the nuances of Rumi’s poetry, showing minimal correlation with human judgment. Crucially, however, even GPT-4o, the most capable model evaluated, failed to perform as high as general knowledge evaluators. This performance ceiling can be attributed to: 1) the inherent complexity of analyzing metaphorical literature, and 2) to the likelihood that these specific poems were absent from the models' pre-training corpus. Consequently, the archaic structures and semantic depth typically present in classical Persian poetry were likely underrepresented in the models’ training distribution which prevents them to learn existing poetic patterns. These findings underscore the significant need for future research dedicated to developing language models capable of genuinely interpreting the rich, metaphorical landscape of literature, including Persian poetry.

Another particularly significant finding emerged from the comparison of the encoder-based models. Pars-BERT was developed by fine-tuning BERT Multilingual on Farsi sentiment datasets; therefore, our initial hypothesis was that its specialized exposure to Farsi would yield superior performance compared to the base model. However, our analysis revealed that Pars-BERT actually performed worse than BERT Multilingual. We attribute this performance deterioration to the domain shift in the training data: since Pars-BERT was fine-tuned primarily on modern Farsi text (e.g., social media and news), its ability to generalize to the archaic linguistic structures of Rumi’s poetry was compromised. This result is critically important, as it demonstrates that fine-tuning with Out-of-Distribution (OOD), even if language-related, data can adversely affect model performance on specialized tasks like classical literature analysis.

Based on the results presented in Figure \ref{fig:3-sentiment_meters}, the meters utilized to create poems with highest sentiments in Rumi’s work are R24 (\foreignlanguage{persian}{فعلاتن فعلاتن فعلاتن فعلاتن}), R34 (\foreignlanguage{persian}{مستفعلن مستفعلن مستفعلن مستفعلن}), C12 (\foreignlanguage{persian}{مفتعلن مفتعلن فاعلن}) , and R41 (\foreignlanguage{persian}{مفتعلن فاعلن مفتعلن فاعلن}). 
In Parvin E'tesami’s poems, the highest sematic scores are exhibited by C8 (\foreignlanguage{persian}{مفاعلن فعلاتن مفاعلن فعلن}), C9 (\foreignlanguage{persian}{مفاعیلن مفاعیلن فعولن}), and C4 (\foreignlanguage{persian}{فعلاتن فعلاتن فعلن}) Meters. As highlighted in these figures, a significant portion of Rumi’s poems, more than 60\% under certain meters, such as C12 (\foreignlanguage{persian}{مفتعلن مفتعلن فاعلن}), exhibit happy sentiment.  In contrast, poems with positive sentiments within a single meter in Parvin E'tesami’s work reached to maximum of 12\%. This observation aligns with the insights of Prof. Mohammad-Reza Shafiei Kadkani in his book Poetry and Music (\foreignlanguage{persian}{موسیقی شعر}) \citep{30-kadkani1965poetry}, where he notes that many of the poems in Rumi's Divan-i Shams are composed with meters that contain repetitive subparts. These short repetitive rhythmic patterns are intentionally created mostly for Sufism (Tasawwuf) gatherings, where the communal recitation of poems was designed to create a sense of enthusiasm and spiritual energy. Therefore, it is understandable to expect that Rumi’s poems exhibit happier and more energetic tone and hence higher sentiment scores compared to Parvin E'tesami’s artwork. However, the primary focus of our research goes beyond confirming this poetic intuition. We aimed to perform a feasibility analysis on LLMs ability to recognizing the complex emotional landscapes embedded within the structure of Persian poems and grasp the sentiments of these poems accurately. As shown in Figure \ref{fig:3e}, between the common meters in Rumi’s and Parvin E'tesami’s poems, Rumi’s poems consistently showed higher averaged sentiment scores; indicating that Rumi utilized these meters to create happier poetic tone compared to Parvin E'tesami. Notably, the average sentiment scores for Parvin E'tesami’s poems under each meter are below 3, indicating that, on average, all of the meters in her work are used to convey predominantly sad sentiments. These findings not only reflect the distinct emotional tones between the two poets but also mark a significant shift in the way we analyze poetry. We believe these results mark the beginning of a new era in the analysis of Persian poetic works, where LLMs make it possible to perform automatic conceptual statistical analysis. While, previously, computer-based systems could only perform exterior structural analyses and semantic and emotional investigations required human intervention.

A comparison of meter usage in Rumi’s versus Parvin E'tesami’s poems can indicate which of the poets utilize the meters in a more artistically diverse way. However, here our aim is not to argue that Rumi has utilized the meters in a more creative manner to compose poems with varied sentiments, as it is well-established among those familiar with Persian poetry that Rumi’s poems are considered one the most significant examples of masterful utilization of meters in Persian poem corpus. Instead, our aim is to determine whether sentiment analysis results extracted using computers, without human intervention, can yield similar conclusions. To address this, we calculated Entropy parameter for each meter. As expected from its formula (Equation \ref{eq:entropy}), given that there are 5 possible sentiment scores in our study, entropy reaches its maximum value when each sentiment score is evenly distributed, i.e., 20\% of the poems have sentiment score of 1, 20\% sentiment score of 2, and so on. This represents the most diverse distribution of sentiments within a meter. Entropy is minimum, when all poems have the same sentiments under a specific meter, for example, when 100\% of the poems have sentiment score of 1, or 100\% of the poems have sentiment score of 5, etc. Notably, given that in our dataset we only have 5 possible numbers for the sentiment scores, entropy changes within the range of $0 \leqq H(X) \lesssim 2.322$. 

As illustrated in Figure \ref{fig:4-entropy_comparison}, the entropy of meter usage by Rumi is significantly higher than Parvin E'tesami, indicating that Rumi utilized meters to express broader range of sentiments. Interestingly the entropy for certain meters such as R47 (\foreignlanguage{persian}{مفتعلن مفتعلن مفتعلن مفتعلن}), C13 (\foreignlanguage{persian}{مفعول فاعلات مفاعیل فاعلن}), and C2 (\foreignlanguage{persian}{فاعلاتن فاعلاتن فاعلن}), are approximately 2.25, close to the maximum possible value for entropy. This may be one of the reasons that Divan-i Shams by Rumi is considered as one of the richest sources and most masterful examples in utilizing meters in Persian poems corpus. In contrast, in Parvin E'tesami’s poems, the maximum value of entropy is observed in C9 (\foreignlanguage{persian}{مفاعیلن مفاعیلن فعولن}) meter and it reaches to approximately 1.75. These results highlighted that analysis of the poems of Rumi and Parvin E'tesami with LLMs successfully indicate superiority of Rumi’s poems in expressing various emotional sentiments using meters.

To further evaluate the emotional depth in Rumi’s poetry, we analyzed the overall variability of sentiments across his poems using standard deviation and polarization parameters. This approach provided another perspective about the value of his artwork. Figure \ref{fig:5a} shows that when we calculate the average sentiment across all of poems in Divan-i Shams, and then measure the deviation of the sentiments in each poem from the average sentiment, the resulting deviation in Rumi’s work is larger compared to Parvin E'tesami’s poems. This indicates that Rumi has expressed a large variety of the emotional sentiment from his average sentiment while most poems of Parvin E'tesami was closer to her average sentiment, and less deviation is observed in her poems. These findings become more significant when we compare the sentiment polarization in Rumi’s poems with those of Parvin E'tesami. We define polarized poems, as those with sentiment of either sad (sentiment score of 1or 3) or happy (sentiment score of 4 or 5) and neutral poems, as those with neutral sentiment (sentiment score of 3). Figure \ref{fig:5b} shows that in total Rumi has more poems with neutral sentiments (sentiment score of 3), however, even though the portion of the polarized poems is less than portion of polarized poems in Parvin E'tesami’s artwork, his masterful usage of the meters led him to compose a wider spectrum of sentiments.

Overall, these analyses demonstrated that LLMs can correctly conduct conceptual comparison between Rumi’s and Parvin E'tesami’s artwork and perform semantic analysis of the Persian poems without requiring direct human interpretations. This is extremely important as it shows the feasibility of utilizing LLMs in investigations of non-English complicated poetic artworks and creates new opportunities in large scale automatic multi-lingual studies in the field of humanities.

\section{Conclusions}

In the present study we employed multiple language models including BERT Multilingual Sentiment Analysis, Pars-BERT, GPT 4o, and GPT 4o-mini models to extract the sentiment of poems from Divan-i Shams by Rumi and Divan-i Ashaar by Parvin E'tesami. Our analysis focused on finding correlations between sentiments and meter of the poems. Our findings led to the following conclusions:

\begin{enumerate}
    \item LLMs exhibited significant performance disparities. While GPT-4o demonstrated the highest resemblance to human scholars (QWK $\sim0.60$), the BERT-based models failed to achieve meaningful correlation. This suggests that large-scale generative models currently outperform smaller encoder-based models in interpreting the nuances of classical Persian literature.
    
    \item Contrary to expectations, ParsBERT (fine-tuned on modern Farsi) performed worse than the base BERT Multilingual model. This indicates that fine-tuning on modern, out-of-distribution text (e.g., news, social media) can deteriorate performance on archaic, metaphorical texts.
    
    \item While models struggled to replicate exact human grading on a 5-point ordinal scale (low exact accuracy), GPT-4o showed strong performance in identifying directional polarity (Positive vs. Negative). Therefore, while LLMs are not yet fully reliable for granular sentiment intensity, they are effective tools for macro-level sentiment classification in classical poetry.
    
    \item Poems composed by Rumi expressed happier sentiments compared to Parvin E'tesami’s poems and this trend remained consistent when poems were grouped and analysed based on their meters.
    
    \item Sentiment analysis results showed that Rumi utilized meters for expressing more variety of the tones and this may be one of the reasons that his work is considered one of the best examples of utilizing meter in a masterful way in Persian poetry corpus.
    
    \item Our results highlighted that the standard deviation of sentiment scores in Rumi’s poems is significantly higher than in Parvin E'tesami’s. This indicates that Parvin E'tesami’s poems follow a more stable and consistent emotional tone, while the sentiments in Rumi’s poems show a larger deviation from the average sentiment score.
    
\end{enumerate}

In summary, while LLMs offer a promising avenue for scalable, automated studies of Persian poetry, they currently face a performance ceiling due to the complexity of archaic metaphors and the lack of representative pre-training data. Future work must focus on bridging the gap between modern training corpora and classical literary heritage to fully unlock the potential of AI in digital humanities.

\section{Generative AI Usage Disclosure}

The authors utilized artificial intelligence tools during the preparation of this manuscript exclusively for grammatical editing and language refinement.


\newpage

\bibliographystyle{unsrtnat}
\bibliography{references}

\end{document}